\documentclass[10pt,twocolumn,letterpaper]{article}

\usepackage{iccv}
\usepackage{times}
\usepackage{epsfig}
\usepackage{graphicx}
\usepackage{amsmath}
\usepackage{amssymb}
\usepackage{booktabs}
\usepackage{makecell}
\usepackage{multirow}
\usepackage{float}
\usepackage{algorithm}
\usepackage{algorithmicx}
\usepackage{algpseudocode}
\usepackage[accsupp]{axessibility}  % Improves PDF readability for those with disabilities.

\newcommand{\RNum}[1]{\uppercase\expandafter{\romannumeral #1\relax}}

\usepackage[pagebackref=true,breaklinks=true,letterpaper=true,colorlinks,bookmarks=false]{hyperref}

\iccvfinalcopy % *** Uncomment this line for the final submission

\def\iccvPaperID{11288} % *** Enter the ICCV Paper ID here

\begin{document}

%%%%%%%%% TITLE
\title{TransFER: Learning Relation-aware Facial Expression Representations \\ with Transformers}

\author{
Fanglei Xue\textsuperscript{1,2 \footnotemark[1]},
Qiangchang Wang\textsuperscript{3 \footnotemark[1] },
Guodong Guo\textsuperscript{4,5,3 \footnotemark[2]} \\
 \textsuperscript{1}University of Chinese Academy of Sciences, Beijing, China\\ 
\textsuperscript{2}Key Laboratory of Space Utilization, Technology and Engineering Center for Space Utilization,\\ Chinese Academy of Sciences, Beijing, China\\
\textsuperscript{3}West Virginia University, Morgantown, USA\\
 \textsuperscript{4}Institute of Deep Learning, Baidu Research, Beijing, China\\
 \textsuperscript{5}National Engineering Laboratory for Deep Learning Technology and Application, Beijing, China\\
 \tt\small xuefanglei19@mails.ucas.ac.cn,  qw0007@mix.wvu.edu, guoguodong01@baidu.com
}

\maketitle
% Remove page # from the first page of camera-ready.
% \ificcvfinal\thispagestyle{empty}\fi

%%%%%%%%% ABSTRACT
\begin{abstract}
Facial expression recognition (FER) has received increasing interest in computer vision. We propose the TransFER model which can learn rich relation-aware local representations. It mainly consists of three components: Multi-Attention Dropping (MAD), ViT-FER, and Multi-head Self-Attention Dropping (MSAD). First, local patches play an important role in distinguishing various expressions, however, few existing works can locate discriminative and diverse local patches. This can cause serious problems when some patches are invisible due to pose variations or viewpoint changes. To address this issue, the MAD is proposed to randomly drop an attention map. Consequently, models are pushed to explore diverse local patches adaptively. Second, to build rich relations between different local patches, the Vision Transformers (ViT) are used in FER, called ViT-FER. Since the global scope is used to reinforce each local patch, a better representation is obtained to boost the FER performance. Thirdly, the multi-head self-attention allows ViT to jointly attend to features from different information subspaces at different positions. Given no explicit guidance, however, multiple self-attentions may extract similar relations. To address this, the MSAD is proposed to randomly drop one self-attention module. As a result, models are forced to learn rich relations among diverse local patches. Our proposed TransFER model outperforms the state-of-the-art methods on several FER benchmarks, showing its effectiveness and usefulness.

\end{abstract}

%%%%%%%%% BODY TEXT
\renewcommand{\thefootnote}{\fnsymbol{footnote}} %将脚注符号设置为fnsymbol类型，即特殊符号表示

\footnotetext[1]{The first two authors contributed equally. This work was done when Fanglei Xue and Qiangchang Wang were interns at IDL, Baidu Research.}
\footnotetext[2]{Corresponding author
}
\renewcommand{\thefootnote}{\arabic{footnote}}

\section{Introduction}

In the past several decades, facial expression recognition (FER) has received increasing interest in the computer vision research community, as it is important to make  computers understand human emotions and interact with humans.

\begin{figure}
\begin{center}
\includegraphics[width=6cm]{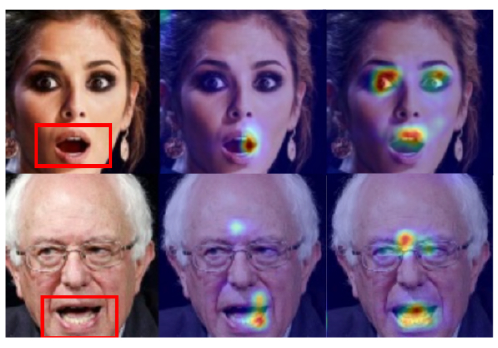}
\end{center}
\caption{Attention visualizations \cite{chefer2020transformer} on two example images: Surprise (Row 1) and Anger (Row 2). Column 1: Original images. Column 2: Attention visualizations of our ViT-FER model. Column 3: Attention visualization of our TransFER model.}
\label{fig_ab}
\end{figure}

Despite it had obtained excellent performance recently, FER is still a challenging task mainly due to two reasons: 1) Large inter-class similarities. Expressions from different classes may only exhibit some minor differences. As illustrated in Fig. \ref{fig_ab}, Surprise (Row 1) and Anger (Row 2) share a similar mouth. Critical clues to distinguish them lie in both eyes and areas between eyes; 2) Small intra-class similarities. Expressions belonging to the same class may have dramatically different appearances, varying with races, genders, ages to cultural backgrounds.

Existing works can be divided into two categories: global-based and local-based approaches. For the former, many loss functions are proposed to enhance the representational ability of features \cite{li2017reliable,farzaneh2021facial}. However, since these methods take global facial images as the input, they may neglect some critical facial regions which would play an important role in distinguishing different expression classes. To overcome this issue, many local-based methods are proposed to learn discriminative features from different facial parts which can be divided into two sub-categories: landmark-based and attention-based approaches. \cite{xie2018facial, happy2014automatic,xie2019deep} extracted features on facial parts which are cropped around landmarks. However, there are several issues: 1) Pre-defined facial crops may not be flexible to describe local details which may vary from different images. This is because important facial parts may appear at different locations, especially for faces with pose variations or viewpoint changes; 

2) Facial landmark detection may be inaccurate or even fail for faces which are affected by various challenging factors, such as strong illumination changes, large pose variations, and heavy occlusions. Therefore, it is necessary to capture important facial parts and suppress useless ones.

To achieve the aforementioned goal, \cite{li2018occlusion,wang2020region} applied attention mechanisms. However, they may have redundant responses around similar facial parts, while neglecting other potentially discriminative parts which would play an important role in FER. This issue is especially serious for faces with occlusions or large pose variations where some facial parts are invisible. Therefore, diverse local representations should be extracted to classify different expressions. Consequently, 
more diverse local patches can contribute even when some patches are invisible. Meanwhile, different local patches can be complementary to each other.
For example, as illustrated in Fig.~\ref{fig_ab}, it is difficult to distinguish between surprise (Row 1) and anger (Row 2) based on the mouth areas only (Column 2). Our TransFER model explores diverse relation-aware facial parts, like eyes (Column 3, Row 1) and areas between the brows (Column 3, Row 2), which help distinguish these different expressions. Thus, the relations among different local patches should be explored in a global scope, highlighting important patches and suppressing the useless.

To achieve the above two goals, we propose the TransFER model to learn diverse relation-aware local representations for FER. First, the Multi-Attention Dropping (MAD) is proposed to randomly drop an attention map. In such a way, models are pushed to explore comprehensive local patches except for the most discriminative ones, focusing on diverse local patches adaptively. This is especially useful if some parts are invisible due to pose variations or occlusions. Second, Vision Transformer (ViT) \cite{dosovitskiy2020image} is adapted to FER, called ViT-FER, to model connections 
among multiple local patches. Since the global scope is used to reinforce each local patch, the complementarity 
among multiple local patches are well explored, boosting the recognition performance. Third, multi-head self-attention allows ViT to jointly attend to features from different information subspaces at different positions. Redundant relations may be built, however, since there is no explicit guidance. To address this, Multi-head Self-Attention Dropping (MSAD) is proposed to randomly drop one self-attention. In such a manner, if a self-attention is dropped, models are forced to learn useful relations from the rest. Consequently, rich relations among different local patches are explored to benefit the FER.

Combining the novel MAD and MSAD modules, we propose the final architecture, termed as TransFER. As illustrated in Fig. \ref{fig_ab}, compared with the ViT-FER baseline (Column 2), the TransFER locates more diverse relation-aware local representations (Column 3), distinguishing these different expressions. It achieves the state-of-the-art performance on several FER benchmarks, showing its effectiveness. The contributions of this work can be summarized as follows:

1. We apply ViT to characterize the relations between different facial parts adaptively, called ViT-FER, showing their effectiveness for FER. To the best of our knowledge, this is the first effort to explore Transformers and investigate the importance of relation-aware local patches for FER.

2. A Multi-head Self-Attention Dropping (MSAD) is introduced to randomly remove self-attention modules, forcing models to learn rich relations between different local patches.

3. An Multi-Attention Dropping (MAD) is designed to erase attention maps, pushing models to extract comprehensive local information from every facial part beyond the most discriminative parts.

4. Experimental results on several challenging datasets show the effectiveness and usefulness of our proposed TransFER model.

\section{Related Work}

\begin{figure*}
\begin{center}
\includegraphics[width=16cm]{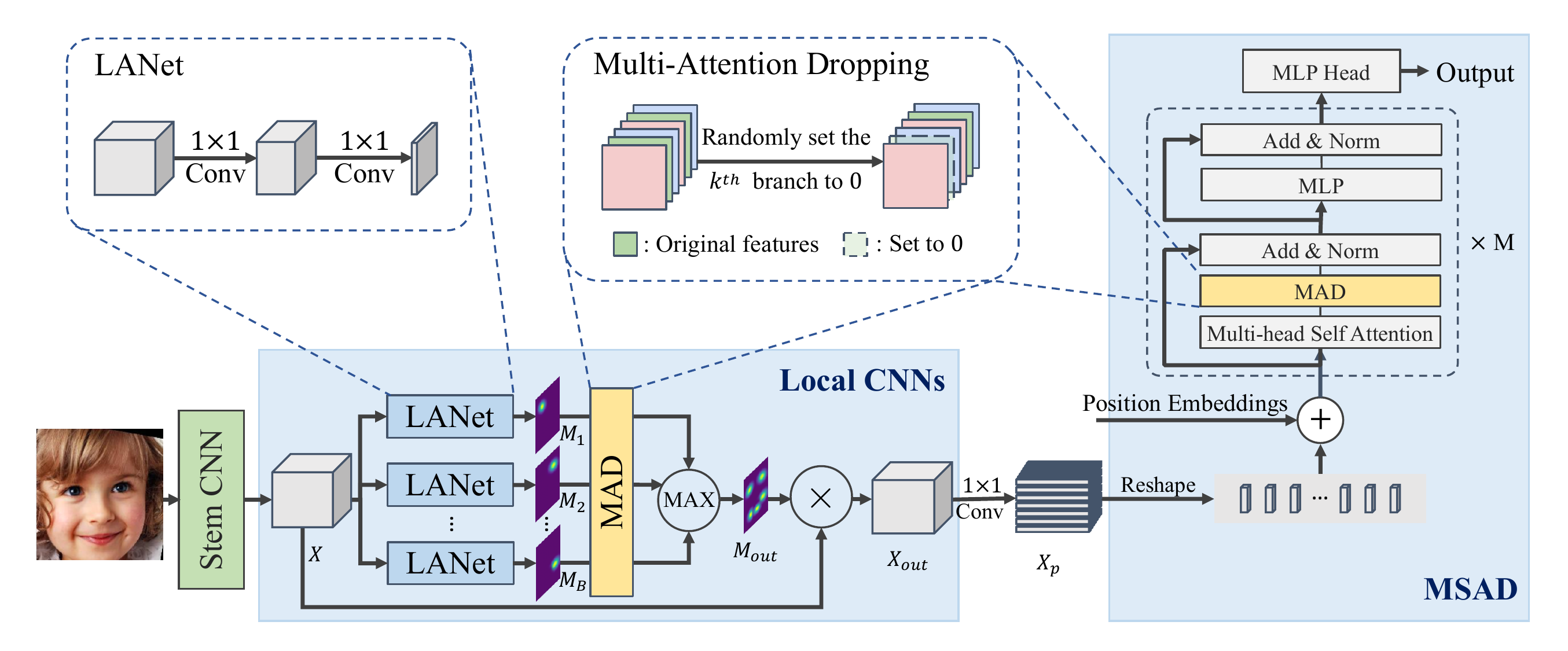}
\end{center}
\caption{The overall architecture of our TransFER model. Firstly, facial images are first fed into a stem CNN to extract feature maps. Secondly, feature maps are then passed through the local CNNs to locate diverse useful feature areas. Thirdly, a $1\times1$ convolution and reshape operations are used to project feature maps to a sequence of feature vectors which can be directly input into MSAD (MAD in a Transformer encoder) where the relationships between these local patches are explored. An MLP Head is attached to generate the final classification result. MAD guides multiple local branches to locate diverse local patches. MSAD pushes multi-head self-attention to explore rich relations among different local patches.}

\label{fig_model}
\end{figure*}

In this section, related work about facial expression recognition, Transformers, and regularization methods are reviewed briefly.

\subsection{Facial Expression Recognition}

Facial expression recognition (FER) has remained as an active research area in the past decades. Traditionally, the hand-crafted features were developed to describe different facial expressions, such as LBP \cite{shan2009facial}, HOG \cite{dalal2005histograms}, and SIFT \cite{ng2003sift}. However, these features lack of generalization ability under some challenging scenarios, such as poor illumination conditions. 

Recently, deep learning has greatly improved the FER research. Loss functions are designed in \cite{li2017reliable,farzaneh2021facial} to enhance the discriminative ability of expression features. Each ROI is weighed in \cite{li2018occlusion} via a gate unit that computes a weight from the region itself. A region attention network is proposed in \cite{wang2020region} to adaptively capture the importance of facial regions for occlusion and pose variant FER. 

Differently, a new adaptive loss is proposed in \cite{li2021adaptively} to re-weight category importance coefficients, alleviating the imbalanced class distribution. Besides, several works \cite{wang2020suppressing,chen2020label} address label uncertainties in FER. In our approach, a mechanism is designed to locate diverse local patches. Besides, relations among different local patches are captured, which is the first attempt for FER, to the best of our knowledge.

\subsection{Transformers in Computer Vision}

Recently, Transformers \cite{vaswani2017attention} are applied to address computer vision problems \cite{carion2020end,dosovitskiy2020image,touvron2020training}.

An end-to-end object detection method reason about the locations of the objects, utilizing the Transformer decoder \cite{carion2020end}. More recently, Vision Transformer (ViT) \cite{dosovitskiy2020image} treats images as a sequence of patches for image classification. Pre-trained on large-scale datasets, it obtained a competitive performance. Without the requirement of large-scale training data, DeiT \cite{touvron2020training} can accelerate training using a teacher-student strategy. In our work, it is the first attempt to explore Transformers for FER, to the best of our knowledge. Besides, this is also the first effort to show the importance of relations among local patches for FER.

\subsection{Regularization Method}
Overfitting is an important issue in deep neural networks. Dropout \cite{srivastava2014dropout} randomly zeroes some of the elements in fully-connected layers, alleviating the overfitting problem. Despite its effectiveness, it is less effective in convolutional operations. This is because features are spatially correlated in CNNs. To address this challenge, Cutout \cite{devries2017improved} is proposed to randomly erase contiguous regions in the input image. DropBlock \cite{ghiasi2018dropblock} further improves the Cutout by applying Cutout at every feature map. We propose the 
MSAD to effectively regularize the Transformers, exploring rich relations among different local patches.

\section{TransFER}

The overall architecture of our approach is shown in Fig~\ref{fig_model}, which mainly consists of the stem CNN, Local CNNs, and Multi-head Self-Attention Dropping (MSAD). The stem CNN is used to extract feature maps.
The IR-50 \cite{deng2019arcface} is adopted here since it has a good generalization.

As mentioned above, due to the small inter-class differences among different emotions, it is highly desired to extract diverse local patches. To achieve this goal, Multi-Attention Dropping (MAD) is devised to randomly drop facial parts. In such a way, multiple local branches in local CNNs are encouraged to locate diverse discriminative local patches. In MSAD, rich relations between different local patches are captured to boost the FER performance. This is achieved by randomly dropping self-attention modules. As a result, multi-head self-attentions are complementary with each other, learning rich useful relations among different local patches. More details are illustrated as follows.

\subsection{Local CNNs}\label{section:localCNN}

As described earlier, given a facial image, our approach first uses a stem CNN to extract feature maps. Then, multiple spatial attentions are used to capture local patches automatically. However, if without a proper guidance, it is not guaranteed that comprehensive discriminative facial parts are located. If models focus on few discriminative facial parts, FER would suffer from performance degradation when these parts are hard to recognize or totally occluded, especially for faces with large pose variations, or strong occlusions. To address this, local CNNs are developed to extract diverse local features which are guided by the MAD.

The framework is shown in Fig.~\ref{fig_model}, mainly consists of three steps, which are detailed as follows.

Firstly, multiple attention maps are generated. Let $X \in R^{h\times w\times c}$ denote the input feature maps where $h, w,$ and $c$ refer to the height, width, and the number of feature maps, respectively. Since LANet \cite{wang2019ls} allows models to automatically locate important face parts, it is used in multiple local branches, as illustrated in Fig.~\ref{fig_model}. It consists of two $1\times1$ convolution layers. The first one outputs $c/r$ feature maps where $r$ is the reduction ratio to reduce the dimension of feature maps, followed by a ReLU layer to enhance the non-linearity. The second layer reduces the feature map number to one and generates an attention map by a Sigmoid function, which is denoted as $M_i$. Suppose there are $B$ LANet branches, then attention maps $[M_1, M_2, ..., M_B]$ are generated where $M_i \in R^{h\times w\times1}$.

Secondly, the MAD forces multiple local branches to explore diverse and useful facial parts, which would be presented in Section \ref{section:MAD}. Generally speaking, it takes several branches of data as input and randomly drops one branch by setting the values in this branch to zeroes (without changing the input shape). As a result, MAD takes $B$ attention maps as input, randomly set one attention map to zeroes, and output $B$ attention maps.

Thirdly, multiple attention maps are aggregated together to generate one attention map. To be specific, an element-wise maximum operation is used to aggregate multiple attention maps. Given a list of feature maps $[M_1, M_2, \ldots, M_B]$, the output $M_{out}$ can be formulated as follows:
\begin{equation}
% \begin{split}
	M_{out}(x, y) = \max\{M_1(x, y), M_2(x, y) \ldots M_B(x,y)\}
% \end{split}
\end{equation}
where $1\leq x \leq w$ and $~1\leq y \leq h$.

Finally, we multiply $M_{out}$ with the original feature map $X$ using an element-wise production. Thus, unimportant areas in the original feature map are suppressed and vise versa. 

To summarize, local CNNs are able to locate diverse local patches. This is achieved by using multiple LANets to locate multiple discriminative areas and aggregate them by a maximum operation, followed by element-wise multiplication with the input feature maps.

\subsection{Multi-Attention Dropping}
\label{section:MAD}

Dropout \cite{srivastava2014dropout} is proposed to prevent neural networks from overfitting. It adapts a feature vector or feature map as input. During the training process, some of the elements of the input are randomly set to zeroes with a probability $p$ using samples from a Bernoulli distribution. If there is more than one channel, each channel would be zeroed out independently. Inspired by this works, a dropout-like operation is developed for the FER task, called Multi-Attention Dropping.

In contrast with the standard Dropout, our proposed MAD adopts a group of feature maps (or vectors) as input and treats every feature map as a whole. As shown in the middle-upper part of Fig.~\ref{fig_model}, during the training process, one feature map is selected from a uniform distribution which is entirely set to zeroes. The drop operation is performed with a probability $p_1$. Dropped feature maps would not be activated in the following layers. Thus, a dropout-like stop-gradient operation is proposed, which can guide local CNNs to explore diverse and discriminative facial parts. As a consequence, well-distributed facial parts can be located, leading to comprehensive local representations to benefit the FER.

\subsection{Multi-head Self-Attention Dropping}

In order to explore the rich relationships among different local features generated by local CNNs, the Multi-head Self-Attention Dropping (MSAD) module is proposed. It mainly consists of a Transformer encoder with MAD injected behind every Multi-head Self Attention module and an MLP classification head like Vision Transformer (ViT) \cite{dosovitskiy2020image}  does. The following are the details:

\textbf{Projection.} After local CNNs, feature maps $X_{out} \in R^{h\times w\times c}$ are generated which contain information about diverse local patches. To capture rich relations among multiple local patches, the Transformer is used which contains multiple encoder blocks. However, since the Transformer is first proposed for NLP tasks and adopts a sequence of 1D feature vectors as input. To adapt the Transformer, a projection module is developed to transform 2D sequence input to 1D.

As illustrated in Fig.~\ref{fig_model}, a $1 \times1$ convolution layer is first applied to the $X_{out}$, projecting to feature maps $X_p \in R^{h\times w\times c_2}$ where the number of channels is denoted as $c_2$. So far, we do not change the height and width relationship between the feature maps $X_p$ and the original image. So, every $c_2 \times 1 $ vector can be considered as a representation of a corresponding patch of the input image. So, we slice the $X_p$ feature maps along the channel dimension and realign them as a sequence of feature vectors $x \in R^{(h \cdot w)\times c_2}$ which can be fed into the Transformer encoder directly. 

Following \cite{dosovitskiy2020image}, the learnable $[class]$ token is also appended to the sequence of input vectors. And standard learnable 1D position embeddings are added to the expanded sequence of vectors to inject position information.

\textbf{Transformer Encoder.}
The Transformer encoder \cite{vaswani2017attention} is composed of a stack of $M$  encoder blocks. Every block is composed of multiple layers of Multi-head Self-Attention (MSA) and Multi-Layer Perceptron (MLP) with skip connections, as shown in the right of Fig. \ref{fig_model}. 
A classification head implemented by a single layer of MLP is attached to perform classification output.

Firstly, the input $x \in R\textsuperscript{N$\times$ d}$ is linearly transformed to queries $q$, keys $k$, and values $v$ as follows:
\begin{equation}
  [q, k, v]=x[w_{q}, w_{k}, w_{v}],
\end{equation}
where $w_{q}, w_{k} \in R\textsuperscript{d$\times$ d\textsubscript{k}}$, $w_{v} \in R\textsuperscript{d$\times$ d\textsubscript{v}}$.

Secondly, the attention weights are computed as follows:
\begin{equation}
   A=Softmax(\frac{qk^T}{\sqrt d_k}).
\end{equation}

Thirdly, a weighted sum over all values is computed as follows:
\begin{equation}
  O=Av.
\end{equation}

The MSA runs self-attention operations $k$ times in parallel and linearly embeds their concatenated outputs to form the final output.

The MLP consists of two fully-connected layers for feature projection and GELU \cite{hendrycks2016gaussian} for non-linearity.

% \textbf{MAD.}
The MSA is designed to embed the projections in their respective space. However, if without an explicit signal, multiple self-attention modules tend to have redundant projections, limiting the representational ability. To address this issue, we utilize Multi-Attention Dropping (MAD) which is presented in Section \ref{section:MAD} to randomly drop one of the attention heads, pushing models to learn comprehensive relations among different local patches.

Suppose there are $k$ SAs in the MSA. One SA module is randomly selected from a uniform distribution and it is set to zeroes with a probability $p_2$.

MAD is performed across different MSA, that is, every sample in the same mini-batch and every MSA in the different block randomly select one SA from their self $k$ SAs in every training iterations. This brings sufficient randomness to the samples and different MSA blocks in Transformer. Similar to Dropout, MAD is only performed during the training time. But during the inference time, unlike Dropout, MAD did not rescale the weights due to the different mechanisms with fully connected layers.

In such a way, models are encouraged to learn useful information since multiple self-attentions are pushed to complement to each other.

In general, more than one SA can be selected and dropped, but by our observation, dropping two or more SAs at the same time did not increase the performance. So for simplicity, we only consider the drop rate as a hyper-parameter and conduct all of our experiments with dropping only one branch in MAD. 

\section{Experiments}

\subsection{Datasets}

RAF-DB \cite{li2017reliable} is a real-world expression dataset. It contains 29,672 real-world facial images which are collected by Flickr's image search API and independently labeled by about 40 trained human workers. In the experiments, the single-label subset provided in RAF-DB is utilized. It contains 15,339 expression images with six basic expressions (happiness, surprise, sadness, anger, disgust, fear) and neutral expression where 12,271 images of them are used in training and the rest are used for testing. The overall accuracy on the test set is reported.

FERPlus \cite{barsoum2016training} is extended from FER2013 \cite{goodfellow2013challenges} which is a large-scale dataset collected by APIs in the Google image search. It contains 28,709 training, 3,589 validation, and 3,589 test images. They relabeled the dataset with ten labelers to eight emotion categories (six basic expressions, plus neutral and contempt). 
The overall accuracy is reported on the test set.

AffectNet \cite{mollahosseini2017affectnet} is the largest publicly available FER dataset so far. It contains about 1M facial images collected by three major search engines where about 420K images are manually annotated. Follow the settings in \cite{li2021adaptively}, we used 280K training images and 3.500 validation images (500 images per category) with seven emotion categories.
The mean class accuracy on the validation set is reported.

\label{section:implementation_details}
\subsection{Implementation Details}

Since RAF-DB and FERPlus datasets provide annotated landmarks, these landmarks are used for face detection and alignment. For FERPlus dataset, the MTCNN \cite{zhang2016joint} is used to detect and align faces. All images are aligned and resized to $112 \times 112$ pixels. Pre-trained on Ms-Celeb-1M \cite{guo2016ms}, the IR-50 \cite{deng2019arcface} is used as the stem CNN where only the first three stages in IR-50 are used. Pre-trained on ImageNet\footnote{The pre-trained weight is downloaded from https://github.com/rwightman/pytorch-image-models/.}, ViT \cite{dosovitskiy2020image} with eight self-attention heads and a stack of $M=8$ identical encoder layers are adopted as the Transformer Encoder.

Due to the class imbalance problem that is widely existed in FER, upsampling the training data is used to balance the class distribution. The drop rates of MAD in local CNNs ($p_1$) and MSAD ($p_2$) are set to 0.6 and 0.3 for RAF-DB and FERPlus, and 0.2 and 0.6 for AffectNet, respectively based on our grid search.

Our TransFER is trained with the SGD optimizer to minimize the cross-entropy loss. We use the momentum of 0.9 and no weight decay, a mini-batch size of 256 in our experiments. During training, data augmentation is utilized on-the-fly including random rotate and crop, random horizontal flip, and random erasing. At test time, we only resize the original image to $112 \times 112$ pixels and feed it to the model directly. For RAF-DB and FERPlus, we train 40 epochs with an initial learning rate of 1e-3 decayed by a factor of 10 at the 15 and 30 epochs. For AffectNet, due to its large number of samples, we train 20K iterations with an initial learning rate of 3e-4 decayed by a factor of 10 at 9.6K and 19.2K iterations. We train our model on two NVIDIA V100 GPU with 32GB RAM.

\subsection{Ablation Studies}

%% +A +B Table 
\begin{table}
\centering
\caption{Evaluation (\%) of local CNNs, MAD, and MSAD on RAF-DB and AffectNet.}
\label{tab:ab:ab}
\begin{tabular}{ccccc}
\toprule
\makecell[c]{Local\\CNNs} & \makecell[c]{MAD} & MSAD & RAF-DB & AffectNet \\ \midrule
% \makecell[c]{Local\\CNNs} & \makecell[c]{A} & B & RAF-DB & AffectNet \\ \hline
           &              &              & 89.93  & 65.63  \\
$\checkmark$ &              &              & 90.03  & 65.74  \\
$\checkmark$ & $\checkmark$ &              & 90.35  & 65.94  \\
%alu$\checkmark$ &              & $\checkmark$ & 89.99  &   \\
$\checkmark$ & $\checkmark$ & $\checkmark$ & 90.91  & 66.23 \\  \bottomrule
\end{tabular}
\end{table}

\begin{table}[]
\centering
\caption{Evaluation (\%) of different output stages of IR-50 on RAF-DB.}
\label{tab:ab:ir-stage}
\begin{tabular}{@{}cccc@{}}
\toprule
Stage   & 2           & 3     & 4     \\ \midrule
Acc.(\%)  & 84.94 & 90.91 & 90.32 \\ \bottomrule
\end{tabular}
\end{table}

\textbf{Effectiveness of the proposed modules.} 
To validate the proposed modules in our TransFER model, an ablation study is designed to investigate the effects of local CNNs, MAD, and MSAD on RAF-DB and AffectNet, as shown in Table~\ref{tab:ab:ab}. To efficiently show results, a tuple $(a, b)$ is used where $a$ and $b$ denote  the performance on RAF-DB and AffectNet, respectively.

The baseline strategy (the first row) means there is no local CNN, no MAD, and no MSAD. The feature maps extracted from the stem CNN are directly fed into the standard Transformer encoder without any guide from MAD or MSAD.
Compared with the baseline, local CNNs slightly improve the performance by $(0.1\%, 0.11\%)$,  but gains significant improvement with the addition of MAD $(0.42\%, 0.30\%)$. It is hypothesized that multiple LANets cannot generate diverse attention maps without extra supervision. MAD achieves this by randomly dropping one LANet branch during the training process, guiding the local CNNs to explore more recognizable feature areas. The MSAD further improves the performance, which achieves the state-of-the-art performance of $(90.91\%, 66.23\%)$, improved by $(0.56\%, 0.29\%)$. We believe that this is due to the multiple self-attentions that are pushed to complement each other and learn comprehensive and useful 
representations.

\textbf{Determination of Stem CNN depth.}
As we know, given a CNN network, deeper layers generate more high-level and semantic information while shallow layers contain more texture and detailed information. For our proposed framework,
we need both semantic information for local CNNs to locate more precise positions and detailed information to fed into MSAD for further extraction.
So we designed this ablation study to determine which stage of IR-50 is the best for the FER task.

Like ResNet-50, IR-50 has four stages, every stage is made up of two convolutional layers and a max pooling layer to quarter the feature map size. Since the input image is in size $112 \times 112$, the output feature size in four stages are $56\times56$, $28\times28$, $14\times14$, $7\times7$, respectively. The feature map size of the first stage out is very large, making too many parameters for the following modules, so we only examine the stages 2 to 4. 

From the results in Table~\ref{tab:ab:ir-stage}, stage 3 achieves the best performance of 90.91\% while stage 4 gives a comparative performance (90.32\%), but with more parameters. Stage 2 performs much worse, only achieves 84.94\%, proving that the LANet is not able to locate well semantic features.

\begin{figure}
\begin{center}
\includegraphics[width=\linewidth]{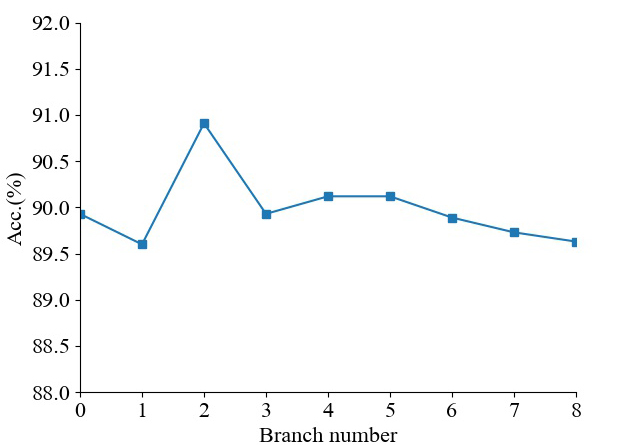}
\end{center}
\caption{The evaluation of branch number ($B$) in Local CNNs on RAF-DB.}
\label{fig:ab:B}
\end{figure}

\textbf{Evaluation of $B$ in local CNNs.}
As we have described in Section \ref{section:localCNN}, $B$ denotes the number of local branches in local CNNs. To explore the impact of branch number $B$, we evaluate the $B$ from 0 to 8 on RAF-DB with other parameters as default. The evaluation results are shown in Figure~\ref{fig:ab:B}. As $B$ increases, the performance first increases and starts to decrease after $B=5$. 

The best performance (90.91\%) is achieved when $B$ is set to 2. Small $B$ makes TransFER hard to locate robust and important feature parts and only achieved 89.60\%. Large $B$ degrades the ability of TransFER, since more branches may fall into a ``collapsing solution" with almost the same outputs. AffectNet is a more difficult dataset, so the best performance is achieved with $B=4$.

\textbf{Evaluation of drop rate in MAD and MSAD.}
To evaluate the impact of drop rates in MAD and MSAD, Experiments of different drop rates are designed on RAF-DB. Denote $p_1$, $p_2$ as the drop rate in MAD and MSAD respectively, they are set to 0.6 and 0.3 by default. 
As shown in Table~\ref{tab:ab:drop-rate}, both small and large $p_1$, $p_2$ values reduce the model performance. 
When $p_2$ is set to 0.3, $p_1$ changes from 0.4 to 0.8, the performance first increases from 89.28\$ to 90.91\%, after that, decreases back to 89.89\%. 

The same phenomenon is observed on $p_2$ when $p_1$ fixed to 0.6. The performance first increases from 89.80\% to 90.91\% when $p_2$ is set to 0.3, and decreases back to 89.24\% as $p_2$ continues to increase. 

There are eight self-attention heads in MSAD while MAD only has two LANet branches. The best $p_2$ value is smaller than the $p_1$ value indicates that self-attention in MSAD can grab important areas more effectively on RAF-DB.

\begin{table}
  \centering
  \caption{Evaluation of different drop rate in MAD ($p_1$) and MSAD ($p_2$) on RAF-DB.}
    \label{tab:ab:drop-rate}
    \begin{tabular}{ccc}
\hline
$p_1$ & $p_2$  & Acc.(\%) \\ \hline
    0.4   & 0.3   & 89.28 \\
    0.5   & 0.3   & 89.24 \\
    0.6   & 0.3   & \textbf{90.91} \\
    0.7   & 0.3   & 89.89 \\
    0.8   & 0.3   & 89.89 \\ \hline
    0.6   & 0.1   & 89.80 \\
    0.6   & 0.2   & 89.83 \\
    0.6   & 0.3   & \textbf{90.91} \\
    0.6   & 0.4   & 89.93 \\
    0.6   & 0.5   & 89.24 \\ \hline
    \end{tabular}%
  \label{tab:addlabel}%
\end{table}%

\textbf{Comparison among MAD, Dropout, Drop Block, and Spatial Dropout.}

First, formally speaking, MAD accepts a set of attention maps as input, randomly selects one and drops the whole selected map. This is the reason why we call it Self-Attention Dropping. In contrast, Dropout \cite{srivastava2014dropout}, Drop Block \cite{ghiasi2018dropblock} and Spatial Dropout \cite{tompsonEfficientObjectLocalization2015} are directly applied to feature maps. Dropout treats all inputs equally and drops them independently, Drop Block drops units in a contiguous region, and Spatial Dropout drops the entire channel. They all perform independently element-wise or channel-wise, which is not suitable for input cases with multiple attention maps. To verify our hypotheses, we perform experiments on RAF-DB and AffectNet with these methods and our proposed MAD. Other hyper-parameters are the default as described in \ref{section:implementation_details}. 

We replace our MAD with these methods and perform a grid search to find the best hyper-parameters. The best result of each method is shown in Tab.~\ref{tab:ab:drops}. Dropout achieves the best performances with dropping rate 0 and 0.1 for RAF-DB and AffectNet, respectively. We also find that,  with a dropping rate of 0.6, the model seems did not work on RAF-DB (39.05\%) but achieves comparative performance on AffectNet (65.51\%). This may because AffectNet contains more training data thus the model can learn from more diverse situations.

The best result of Drop Block is achieved with a dropping rate of 0.3, and the block size is 7 and 9, respectively on RAF-DB and AffectNet. The best dropping rate for Spatial Dropout is 0.2 for both two datasets. Our MAD achieves the best performances, we believe this is because other methods perform dropping independently channel-wise. This works for feature maps because the channel number is big, but not suitable for attention maps with few branches in our case.

\begin{table}[]
\centering
% \footnotesize
\caption{Comparison among our MAD, Dropout, Drop Block and Spatical Dropout.}
\label{tab:ab:drops}
\begin{tabular}{c|cccc}
\toprule
Dataset    & \textbf{MAD}   &
\makecell[c]{Dropout} &
\makecell[c]{Drop\\Block} &
\makecell[c]{Spatial\\Dropout} \\ \midrule
RAF-DB  & \textbf{90.91\%} & 90.35 & 90.25\%     & 89.99\%                      \\
AffectNet & \textbf{66.23\%} & 66.06 & 66.03\%      & 65.54\%            \\ \bottomrule
\end{tabular}
% \vspace{-22pt}
\end{table}

% \begin{table}
% \centering
% \caption{Performance comparison (\%) between Dropout and MAD.}
% \label{tab:ab:dropout}
% \begin{tabular}{c|ccc} 
% \toprule
% \multicolumn{1}{l|}{Type} & \multicolumn{1}{l}{Drop rate} & \multicolumn{1}{l}{RAF-DB} & \multicolumn{1}{l}{AffectNet}  \\ 
% \midrule
% \multirow{7}{*}{Dropout}  & 0    & 90.35    & 65.94   \\
%                           & 0.1    & 89.60    & 66.06   \\
%                           & 0.2    & 89.28   & 65.77            \\
%                           & 0.3    & 89.99   & 65.66            \\
%                           & 0.4    & 89.08   & 65.57            \\
%                           & 0.5    & 80.51   & 65.23            \\ 
%                           & 0.6    & 39.05   & 65.51            \\ 
% \hline
% \multirow{2}{*}{MAD}      & 0.3    & \textbf{90.91} & -                \\
%                           & 0.6    & -       & \textbf{66.23}   \\
% \bottomrule
% \end{tabular}
% \end{table}

%% Compare with SOTA
\begin{table}[H]
\centering
\caption{Performance comparison (\%) with the state-of-the-art methods on RAF-DB and AffectNet.}
\label{tab:sota:rafaffect}
\begin{tabular}{ccc}
\hline
Method                               & RAF-DB & AffectNet \\ \hline
DLP-CNN \cite{li2017reliable}        & 80.89  & 54.47     \\
gACNN \cite{li2018occlusion}         & 85.07  & 58.78     \\
IPA2LT \cite{zeng2018facial}         & 86.77  & 55.71     \\
RAN \cite{wang2020region}            & 86.90  & 52.97     \\
CovPool \cite{acharya2018covariance} & 87.00  & -         \\
SCN \cite{wang2020suppressing}       & 87.03  & 60.23     \\
DACL \cite{farzaneh2021facial}       & 87.78  & 65.20     \\
KTN \cite{li2021adaptively}          & 88.07  & 63.97     \\ \hline
\textbf{TransFER (Ours)} & \textbf{90.91} & \textbf{66.23} \\ \hline
\end{tabular}
\end{table}

\subsection{Comparison with the State of the Art}
Table~\ref{tab:sota:rafaffect} compares our best results to the state-of-the-art methods on RAF-DB and AffectNet. RAF-DB is the latest facial expression dataset, and to our  best knowledge, our proposed TransFER is the first model to achieve accuracy over 90\% on this dataset, which is 2.84\% better than KTN \cite{li2021adaptively}, the best result reported before. AffectNet is the largest dataset of facial expressions, a very challenging dataset. KTN \cite{li2021adaptively} achieved the second-best performance in RAF-DB which is 1.23\% lower than the best result reported previously on AffectNet. Our proposed approach outperforms the previous best result (DACL) by 1.03\%.

Tabel~\ref{tab:sota:ferplus} compares the performance of our TransFER with the state-of-the-art methods on FERPlus. It can be seen that our method achieves the best accuracy of 90.83\%. It is noting that both SCN \cite{wang2020suppressing} and KTN \cite{li2021adaptively} achieve that reported performance by applying trivial loss functions, while we achieve better performance with the standard CE loss only.

\begin{table}[]
\centering
\caption{Performance comparison (\%) with the state-of-the-art methods on FERPlus.}
\label{tab:sota:ferplus}
\begin{tabular}{ccc}
\hline
Method                               & FERPlus \\ \hline
PLD \cite{barsoum2016training}       & 85.10     \\
RAN \cite{wang2020region}            & 88.55     \\
SeNet50 \cite{albanie2018emotion}    & 88.80     \\
RAN-VGG16 \cite{wang2020region}      & 89.16     \\
SCN \cite{wang2020suppressing}       & 89.35     \\
KTN \cite{li2021adaptively}          & 90.49     \\ \hline
\textbf{TransFER (Ours)}     & \textbf{90.83}  \\ \hline
\end{tabular}
\end{table}

\subsection{Attention Visualization}
To further investigate the effectiveness of our approach, we employ the method \cite{chefer2020transformer} to visualize the attention maps generated by our TransFER. To be specific, we first resize the visualization attention maps to the same size as the input images and visualize attention maps through COLORMAP\_JET color mapping to the original image. 

\begin{figure}
\begin{center}
\includegraphics[width=\linewidth]{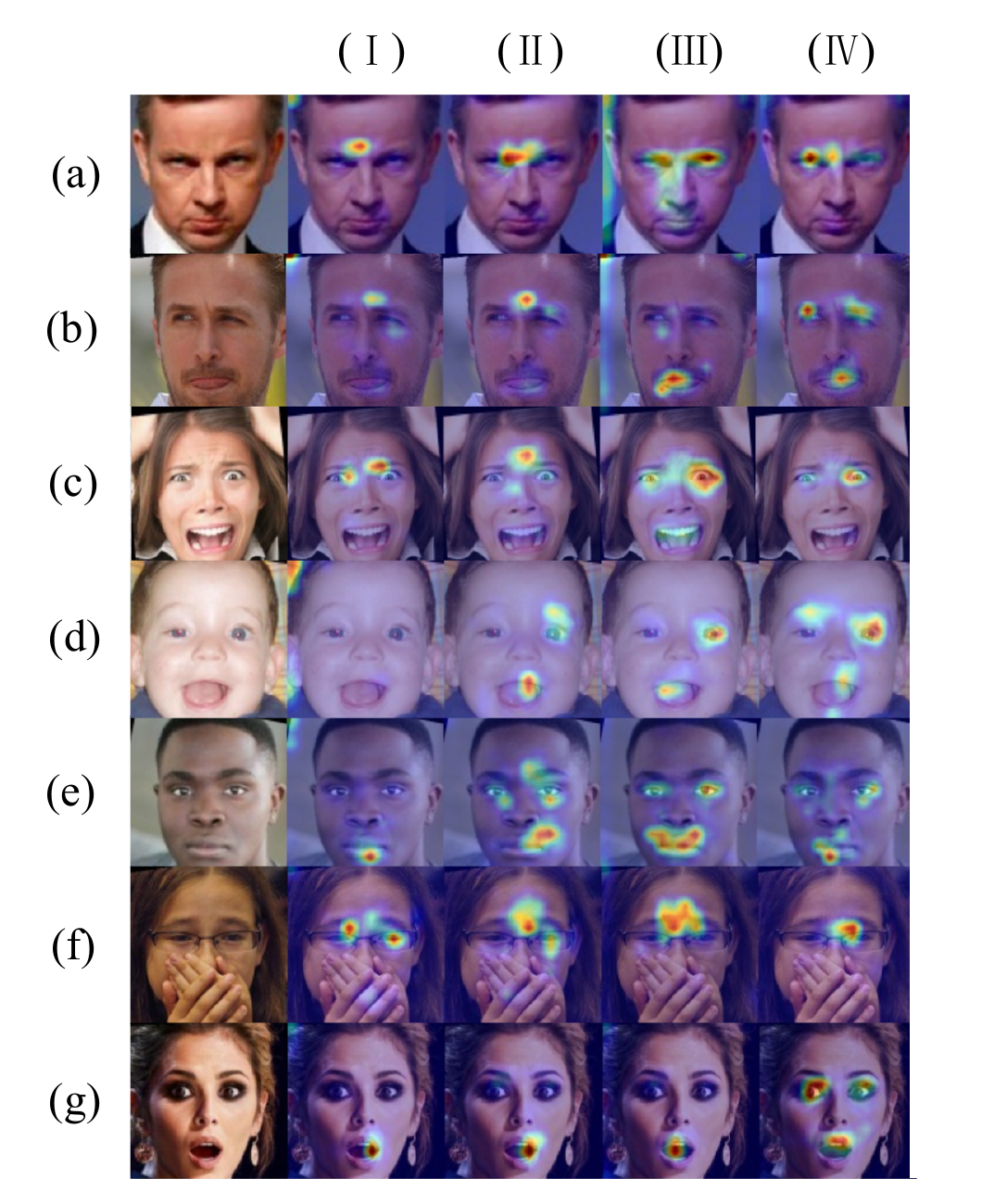}
\end{center}
\caption{Attention visualization \cite{chefer2020transformer} of different expressions on some example face images from AffectNet dataset. (a) - (g) denote anger, disgust, fear, happiness, neutral, sadness, and surprise separately. (\RNum{1}) - (\RNum{4}) denote four training strategies in Tab.~\ref{tab:ab:ab}, (\RNum{1}) denotes the baseline strategy, (\RNum{2}) denote training with local CNNs but without MAD, (\RNum{3}) denotes training with local CNNs and MAD, and (\RNum{4}) denotes our proposed TransFER, training with local CNNs and MSAD. After applying MAD and MSAD, the whole framework can focus on more discriminative facial areas. 
}
\label{fig:vis:class}
\end{figure}

Fig.~\ref{fig:vis:class} shows the attention maps of different emotions in AffectNet. The figure has seven rows, each row shows one of the seven categories of expression. From top to bottom, the categories are anger, disgust, fear, happiness, neutral, sadness, and surprise. The first column shows the original aligned facial images, and the second to fifth columns show the results of four training strategies which have been listed in Table~\ref{tab:ab:ab}: (\RNum{1}) the baseline strategy; (\RNum{2}) adds multiple LANets to generate multiple attention maps but without MAD to guide;  (\RNum{3}) has both multiple LANets and MAD; (\RNum{4}) the whole architecure, including multiple LANets, MAD, and MSAD.

Firstly, comparing different columns (training strategies): local CNNs (\RNum{2}) can locate more potential interesting areas compared with the baseline (\RNum{1}), and MAD in local CNNs (\RNum{3}) and MSAD (\RNum{4}) enhanced these candidate areas by exploring more interesting areas (\eg (a), (b) in (\RNum{3}) ) and constraining less dividing areas (\eg (c) in (\RNum{4}) ).

Secondly, comparing different rows (emotions): It is generally assumed that mouth, nose, and eyes are the most useful regions to distinguish different emotions. But as  discussed in \cite{li2021adaptively}, due to the high similarity of different emotions, these areas may be very similar even for different emotions. For example, fear (c), happiness (d), and surprise (g) often have an open mouth, so it's more important to explore other facial areas to discriminate against different emotions. Our proposed MSAD solves this problem by constraining the activation of the mouth area in (c) (\RNum{4}) and exploring other useful areas in (g) (\RNum{4}), compared to (\RNum{3}).

\section{Conclusion}
We have proposed a new architecture based on the  Transformer for the FER task, called TransFER, which can learn rich, diverse relation-aware local representations. Firstly, a Multi-Attention Dropping (MAD) has been proposed to guide local CNNs to generate diverse local patches, making models robust to pose variations or occlusions. Secondly, the ViT-FER is applied to build rich connections upon multiple local patches where important facial parts are assigned with higher weights and useless ones are assigned smaller weights. Thirdly, the MSAD has been proposed to explore more rich relations among diverse facial parts. To the best of our knowledge, this is the first work to utilize the Transformers for the FER task. Extensive experiments on three public FER datasets demonstrated that our approach outperforms the state-of-the-art methods.

{\small
\bibliographystyle{ieee_fullname}
\bibliography{main}

\begin{thebibliography}{10}\itemsep=-1pt

\bibitem{acharya2018covariance}
Dinesh Acharya, Zhiwu Huang, Danda Pani~Paudel, and Luc Van~Gool.
\newblock Covariance pooling for facial expression recognition.
\newblock In {\em Proceedings of the IEEE Conference on Computer Vision and
  Pattern Recognition Workshops}, pages 367--374, 2018.

\bibitem{albanie2018emotion}
Samuel Albanie, Arsha Nagrani, Andrea Vedaldi, and Andrew Zisserman.
\newblock Emotion recognition in speech using cross-modal transfer in the wild.
\newblock In {\em Proceedings of the 26th ACM international conference on
  Multimedia}, pages 292--301, 2018.

\bibitem{barsoum2016training}
Emad Barsoum, Cha Zhang, Cristian~Canton Ferrer, and Zhengyou Zhang.
\newblock Training deep networks for facial expression recognition with
  crowd-sourced label distribution.
\newblock In {\em Proceedings of the 18th ACM International Conference on
  Multimodal Interaction}, pages 279--283, 2016.

\bibitem{carion2020end}
Nicolas Carion, Francisco Massa, Gabriel Synnaeve, Nicolas Usunier, Alexander
  Kirillov, and Sergey Zagoruyko.
\newblock End-to-end object detection with transformers.
\newblock In {\em European Conference on Computer Vision}, pages 213--229.
  Springer, 2020.

\bibitem{chefer2020transformer}
Hila Chefer, Shir Gur, and Lior Wolf.
\newblock Transformer interpretability beyond attention visualization.
\newblock {\em arXiv preprint arXiv:2012.09838}, 2020.

\bibitem{chen2020label}
Shikai Chen, Jianfeng Wang, Yuedong Chen, Zhongchao Shi, Xin Geng, and Yong
  Rui.
\newblock Label distribution learning on auxiliary label space graphs for
  facial expression recognition.
\newblock In {\em Proceedings of the IEEE/CVF Conference on Computer Vision and
  Pattern Recognition}, pages 13984--13993, 2020.

\bibitem{dalal2005histograms}
Navneet Dalal and Bill Triggs.
\newblock Histograms of oriented gradients for human detection.
\newblock In {\em 2005 IEEE computer society conference on computer vision and
  pattern recognition (CVPR'05)}, volume~1, pages 886--893. Ieee, 2005.

\bibitem{deng2019arcface}
Jiankang Deng, Jia Guo, Niannan Xue, and Stefanos Zafeiriou.
\newblock Arcface: Additive angular margin loss for deep face recognition.
\newblock In {\em Proceedings of the IEEE/CVF Conference on Computer Vision and
  Pattern Recognition}, pages 4690--4699, 2019.

\bibitem{devries2017improved}
Terrance DeVries and Graham~W Taylor.
\newblock Improved regularization of convolutional neural networks with cutout.
\newblock {\em arXiv preprint arXiv:1708.04552}, 2017.

\bibitem{dosovitskiy2020image}
Alexey Dosovitskiy, Lucas Beyer, Alexander Kolesnikov, Dirk Weissenborn,
  Xiaohua Zhai, Thomas Unterthiner, Mostafa Dehghani, Matthias Minderer, Georg
  Heigold, Sylvain Gelly, et~al.
\newblock An image is worth 16x16 words: Transformers for image recognition at
  scale.
\newblock {\em arXiv preprint arXiv:2010.11929}, 2020.

\bibitem{farzaneh2021facial}
Amir~Hossein Farzaneh and Xiaojun Qi.
\newblock Facial expression recognition in the wild via deep attentive center
  loss.
\newblock In {\em Proceedings of the IEEE/CVF Winter Conference on Applications
  of Computer Vision}, pages 2402--2411, 2021.

\bibitem{ghiasi2018dropblock}
Golnaz Ghiasi, Tsung-Yi Lin, and Quoc~V Le.
\newblock Dropblock: A regularization method for convolutional networks.
\newblock {\em arXiv preprint arXiv:1810.12890}, 2018.

\bibitem{goodfellow2013challenges}
Ian~J Goodfellow, Dumitru Erhan, Pierre~Luc Carrier, Aaron Courville, Mehdi
  Mirza, Ben Hamner, Will Cukierski, Yichuan Tang, David Thaler, Dong-Hyun Lee,
  et~al.
\newblock Challenges in representation learning: A report on three machine
  learning contests.
\newblock In {\em International conference on neural information processing},
  pages 117--124. Springer, 2013.

\bibitem{guo2016ms}
Yandong Guo, Lei Zhang, Yuxiao Hu, Xiaodong He, and Jianfeng Gao.
\newblock Ms-celeb-1m: A dataset and benchmark for large-scale face
  recognition.
\newblock In {\em European conference on computer vision}, pages 87--102.
  Springer, 2016.

\bibitem{happy2014automatic}
SL Happy and Aurobinda Routray.
\newblock Automatic facial expression recognition using features of salient
  facial patches.
\newblock {\em IEEE transactions on Affective Computing}, 6(1):1--12, 2014.

\bibitem{hendrycks2016gaussian}
Dan Hendrycks and Kevin Gimpel.
\newblock Gaussian error linear units (gelus).
\newblock {\em arXiv preprint arXiv:1606.08415}, 2016.

\bibitem{li2021adaptively}
Hangyu Li, Nannan Wang, Xinpeng Ding, Xi Yang, and Xinbo Gao.
\newblock Adaptively learning facial expression representation via cf labels
  and distillation.
\newblock {\em IEEE Transactions on Image Processing}, 30:2016--2028, 2021.

\bibitem{li2017reliable}
Shan Li, Weihong Deng, and JunPing Du.
\newblock Reliable crowdsourcing and deep locality-preserving learning for
  expression recognition in the wild.
\newblock In {\em Proceedings of the IEEE conference on computer vision and
  pattern recognition}, pages 2852--2861, 2017.

\bibitem{li2018occlusion}
Yong Li, Jiabei Zeng, Shiguang Shan, and Xilin Chen.
\newblock Occlusion aware facial expression recognition using cnn with
  attention mechanism.
\newblock {\em IEEE Transactions on Image Processing}, 28(5):2439--2450, 2018.

\bibitem{mollahosseini2017affectnet}
Ali Mollahosseini, Behzad Hasani, and Mohammad~H Mahoor.
\newblock Affectnet: A database for facial expression, valence, and arousal
  computing in the wild.
\newblock {\em IEEE Transactions on Affective Computing}, 10(1):18--31, 2017.

\bibitem{ng2003sift}
Pauline~C Ng and Steven Henikoff.
\newblock Sift: Predicting amino acid changes that affect protein function.
\newblock {\em Nucleic acids research}, 31(13):3812--3814, 2003.

\bibitem{shan2009facial}
Caifeng Shan, Shaogang Gong, and Peter~W McOwan.
\newblock Facial expression recognition based on local binary patterns: A
  comprehensive study.
\newblock {\em Image and vision Computing}, 27(6):803--816, 2009.

\bibitem{srivastava2014dropout}
Nitish Srivastava, Geoffrey Hinton, Alex Krizhevsky, Ilya Sutskever, and Ruslan
  Salakhutdinov.
\newblock Dropout: a simple way to prevent neural networks from overfitting.
\newblock {\em The journal of machine learning research}, 15(1):1929--1958,
  2014.

\bibitem{tompsonEfficientObjectLocalization2015}
Jonathan Tompson, Ross Goroshin, Arjun Jain, Yann LeCun, and Christoph Bregler.
\newblock Efficient object localization using {{Convolutional Networks}}.
\newblock In {\em Proceedings of the {{IEEE Computer Society Conference}} on
  {{Computer Vision}} and {{Pattern Recognition}}}, volume 07-12-June, pages
  648--656. {IEEE Computer Society}, Oct. 2015.

\bibitem{touvron2020training}
Hugo Touvron, Matthieu Cord, Matthijs Douze, Francisco Massa, Alexandre
  Sablayrolles, and Herv{\'e} J{\'e}gou.
\newblock Training data-efficient image transformers \& distillation through
  attention.
\newblock {\em arXiv preprint arXiv:2012.12877}, 2020.

\bibitem{vaswani2017attention}
Ashish Vaswani, Noam Shazeer, Niki Parmar, Jakob Uszkoreit, Llion Jones,
  Aidan~N Gomez, Lukasz Kaiser, and Illia Polosukhin.
\newblock Attention is all you need.
\newblock {\em arXiv preprint arXiv:1706.03762}, 2017.

\bibitem{wang2020suppressing}
Kai Wang, Xiaojiang Peng, Jianfei Yang, Shijian Lu, and Yu Qiao.
\newblock Suppressing uncertainties for large-scale facial expression
  recognition.
\newblock In {\em Proceedings of the IEEE/CVF Conference on Computer Vision and
  Pattern Recognition}, pages 6897--6906, 2020.

\bibitem{wang2020region}
Kai Wang, Xiaojiang Peng, Jianfei Yang, Debin Meng, and Yu Qiao.
\newblock Region attention networks for pose and occlusion robust facial
  expression recognition.
\newblock {\em IEEE Transactions on Image Processing}, 29:4057--4069, 2020.

\bibitem{wang2019ls}
Qiangchang Wang and Guodong Guo.
\newblock Ls-cnn: Characterizing local patches at multiple scales for face
  recognition.
\newblock {\em IEEE Transactions on Information Forensics and Security},
  15:1640--1653, 2019.

\bibitem{xie2018facial}
Siyue Xie and Haifeng Hu.
\newblock Facial expression recognition using hierarchical features with deep
  comprehensive multipatches aggregation convolutional neural networks.
\newblock {\em IEEE Transactions on Multimedia}, 21(1):211--220, 2018.

\bibitem{xie2019deep}
Siyue Xie, Haifeng Hu, and Yongbo Wu.
\newblock Deep multi-path convolutional neural network joint with salient
  region attention for facial expression recognition.
\newblock {\em Pattern Recognition}, 92:177--191, 2019.

\bibitem{zeng2018facial}
Jiabei Zeng, Shiguang Shan, and Xilin Chen.
\newblock Facial expression recognition with inconsistently annotated datasets.
\newblock In {\em Proceedings of the European conference on computer vision
  (ECCV)}, pages 222--237, 2018.

\bibitem{zhang2016joint}
Kaipeng Zhang, Zhanpeng Zhang, Zhifeng Li, and Yu Qiao.
\newblock Joint face detection and alignment using multitask cascaded
  convolutional networks.
\newblock {\em IEEE Signal Processing Letters}, 23(10):1499--1503, 2016.

\end{thebibliography}
}

\end{document}